# Cloud Computing framework for Computer Vision Research:

## *An Introduction*

*Yu Zhou*

*Beijing Jiaotong University*

*Yu.Zhou2020@gmail.com*

## 1. Abstract

Cloud computing offers the potential to help scientists to process massive number of computing resources often required in machine learning application such as computer vision problems. This proposal would like to show that which benefits can be obtained from cloud in order to help medical image analysis users (including scientists, clinicians, and research institutes). As security and privacy of algorithms are important for most of algorithms' inventors, these algorithms can be hidden in a cloud to allow the users to use the algorithms as a package without any access to see/change their inside. In another word, in the user part, users send their images to the cloud and configure the algorithm via an interface. In the cloud part, the algorithms are applied to this image and the results are returned back to the user.

My proposal has two parts: (1) investigate the potential of cloud computing for computer vision problems and (2) study the components of a proposed cloud-based framework for medical image analysis application and develop them (depending on the length of the internship). The investigation part will involve a study on several aspects of the problem including security, usability (for medical end users of the service), appropriate programming abstractions for vision problems, scalability and resource requirements. In the second part of this proposal I am going to thoroughly study of the proposed framework components and their relations and develop them. The proposed cloud-based framework includes an integrated environment to enable scientists and clinicians to access to the previous and current medical image analysis algorithms using a handful user interface without any access to the algorithm codes and procedures.

## 2. Motivation to unify the Medical Image Analysis resources

In medical image analysis, a cloud can be defined as a framework between users/customers and resource providers. Unifying these resources (including existing algorithms, hardware, images and so on) in a cloud is important due to several reasons. Four major reasons are:
- When a researcher addresses a new challenge for the analysis of medical images he/she must consider existing solutions to determine if they are suitable or not. Generally, the re-implementation, when the software is not available for that solution, or the re-use of a solution might be limited by various instances. For instance, implementation details are not enough or the researchers are not familiar with programming languages or even solutions are not compatible with their own development environments. Medical image analysis (MIA) researchers spend around 30% of their research time for implementing and evaluating existing solutions, which represents a significant amount of time. Moreover, the evaluation and validation of the new software solutions is limited by the way that these applications are deployed.
- User interfaces is a big challenge for MIA researchers as they need to spend additional time to define a suitable interface for visual input data and present the results when advance visualization is needed. MIA researchers are often concerned about proving concepts and learning new programming languages, and image and visualization techniques sometimes is not an option.
- The generation of ground truth outlined shapes is relevant for the validation of image segmentation algorithms. Ground truth, in the image processing scenario, consists in examples of target shapes that image segmentation should ideally provide as results. These shapes are generated by either radiologists or MIA researchers by means of manual segmentation, which is often, performed using the mouse. This could be both tedious and time consuming.
- For some researchers sharing information is quite a challenge as they believe their research is threatened when their data and algorithms are used by someone. Even though the creation of data repositories along research centres and universities has been accepted as a way to preserve and disseminate research data, sharing algorithms and software solutions is still a challenge.

By unifying all image analysis resources in one environment, cloud computing helps the medical image analysis researchers to test their new algorithms, visualize the results and compare with other algorithms by applying on standard ground truth images without spending time and energy. Also, as users have independent access to remote computing services in a cloud computing framework, then sharing resources becomes safe and secure.

## 3. Proposed Framework

Due to the massive number of programming languages and image processing toolkits, simple sharing of resources as plug and play functionality is difficult. While the proposed cloud computing framework in Figure.1 allows a platform to have an independent access to remote computing services, its web services allow end-users to fully interact with data, information requests as well as applications with a low level of user interaction. In this proposal, I am going to study different components of this framework and develop them using existing software solutions, imaging toolkits and Microsoft technologies. For instance, Visual Studio can be utilized as the development environment, SQL server as the database manager, and Microsoft Workflow foundation as the workflow engine.
A number of work packages have to be studied/developed for this framework:
- Evaluate existing software environment solutions.

- Design standard datasets, metadata and web services.
- Design workflow orchestration and enactment, for example by Windows Workflow Foundation.
- Incorporate existing imaging and visualisation toolkits.
- Design mechanisms and tools to enhance software usability for clinicians.

A number of challenges have to be addressed. The adaptation of existing software will require an analysis and may require significant time. Also, merging toolkits with different languages (such as Matlab, Java and C++) and their communication will need a deep study. The design and development of suitable user interfaces for clinicians is another challenge. The idea is to use multi-touch technology to improve the interaction with image analysis applications as well as the learning experience. Finally, contributions not only should meet the quality but also facilitate the way researchers do it. The design of mechanisms to provide such facility must be analysed and evaluated.

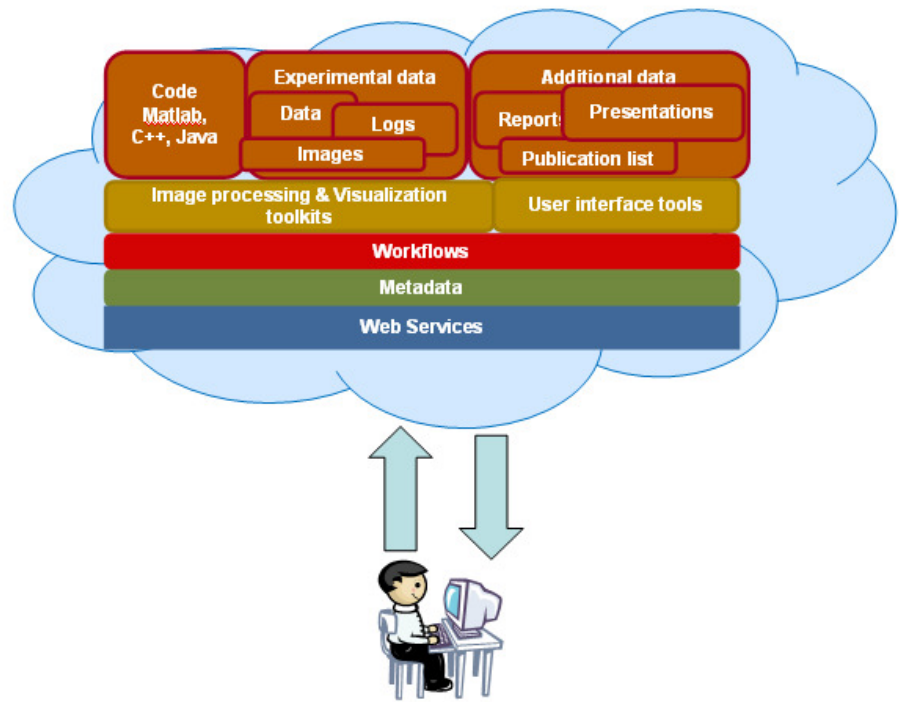

*Figure.1 The proposed framework for medical image analysis application on cloud*